# Evaluation of Distance Measures for Feature based Image Registration using AlexNet

K.Kavitha[1], B. Thirumala Rao[3]
Computer Science and Engineering
K L University
Guntur, A. P., India

B. Sandhya[2]
Computer Science and Engineering
MVSR Engineering College
Hyderabad, India

*Abstract*—Image registration is a classic problem of computer vision with several applications across areas like defence, remote sensing, medicine etc. Feature based image registration methods traditionally used hand-crafted feature extraction algorithms, which detect key points in an image and describe them using a region around the point. Such features are matched using a threshold either on distances or ratio of distances computed between the feature descriptors. Evolution of deep learning, in particular convolution neural networks, has enabled researchers to address several problems of vision such as recognition, tracking, localization etc. Outputs of convolution layers or fully connected layers of CNN which has been trained for applications like visual recognition are proved to be effective when used as features in other applications such as retrieval. In this work, a deep CNN, AlexNet, is used in the place of handcrafted features for feature extraction in the first stage of image registration. However, there is a need to identify a suitable distance measure and a matching method for effective results. Several distance metrics have been evaluated in the framework of nearest neighbour and nearest neighbour ratio matching methods using benchmark dataset. Evaluation is done by comparing matching and registration performance using metrics computed from ground truth.

## I. INTRODUCTION

Every image is identified with its unique and discriminating features which are used as interesting points. Features of an image can be edges, blobs, corners and contours of the objects. Feature extraction forms an integral part of several vision-based applications such as image matching, object tracking, classification etc. The aim of image matching is to map the feature points from one image to corresponding points of another image of the same scene. Matches have to be robust to handle the distortions caused due to noise, changing illumination, reflection, area of projection etc. Image matching is fundamental to image registration which aims at computing the spatial transformation between two images.

Conventional image registration methods are broadly divided into area based and feature based methods [1]. Area based approaches find transformation by optimizing an objective function which is defined based on the error between the similarities of pixel properties of images. Similarity measures such as normalized cross correlation, mutual information are commonly used. Feature based approaches rely on computing similarity between descriptors computed around key points. Several handcrafted feature detection and description algorithms such SIFT, SURF, MSER etc. [2, 3, 4] have been used for the purpose. Spatial transformation between the images is estimated from the matched key points using RANSAC [5]. Performance of feature-based image matching depends to a great extent on the suitability of feature detection and descriptor towards kind of images and deformation [6, 7].

Deep learning is gaining importance in the field of machine learning and is able to provide solutions to several issues of computer vision such as image classification [8]. Among the several deep network architectures such as Deep Belief Network, Convolution Neural Network (CNN), Deep stacking Network, DBoM, Deep Q-Network etc. CNN is gaining attention in the field of imaging. CNN's are formed using multiple layers of convolution, activation, pooling arranged in a hierarchy. The output of these layers forms as input to layers which are typically designed to serve applications such as recognition, tracking, localization etc. The effectiveness of CNN's is because feature extraction process is part of training unlike classical approaches where handcrafted features are used to train only the learning algorithm.

CNN's enable to extract features of an image at different granularities; beginning layers extract basic features, such as lines, borders, and corners, while next level layers exhibit higher features, such as object portions, parts, or the whole object. Outputs from activation layers prior to fully connected layers are being used as features descriptors across different applications. Convolution neural network models like LeNet, AlexNet, VGGNet, ZFNet, GoogleNet, ResNet are being increasingly used as global features, by computing the vector for entire image, in retrieval kind of applications [9, 10, 11, 8, 12]. More recently activation outputs of pretrained CNN's are being used as local feature descriptors. Key points are detected in an image and region around key point is given as input to a CNN to get the feature vector. This paper explores the use of AlexNet features for image registration application.

Section2 describes about the background survey conducted on distance measures used in various applications and use of CNNs, Image registration approach adopted is discussed in Section 3. Section 4 explains data and our approach towards implementation, Section 5 presents evaluation and results, conclusion is part of Section 6.

## II. BACKGROUND WORK

Convolution Neural Networks (CNN's) are being successfully applied in solving several problems of computer vision and natural language processing. First attempt on CNN, LeNet [13], revealed outstanding results in document recognition. Graph Transformer Network with CNN has been used to classify high-dimensional patterns of handwritten characters with more flexibility. Network framed by Alex Krizhevsky, et al. AlexNet, has been trained and tested for classification of high-resolution images, with results better than conventional feature extraction methods like SIFT. In large scale visual recognition experimentation, CNN has shown exceptional performance with VGGNet[10], GoogleNet[14] and many more [8, 12].

CNN eliminates the need of manual feature extraction methods, as the features are extracted directly from images which are learned while network in trained. CNN models which have been trained for applications like visual recognition have been used as feature extractors for several other applications. For example, activations of fc6 layer of AlexNet are proved to improve vehicle image detection and classification [15]. Recently AlexNet features are used in wound tissue analysis and improved version applied for scene classification [16]. ImageNet features are used in classification of earth observation [17].

CNN features are being effectively used in other typical computer vision problems. For example, scene classification based on classical AlexNet extracted features and used along with SVM and regression model [18]. Image retrieval based on AlexNet fc8 layer features admit good recall rate and fusing the fc6 features of LeNet and fc8 features of AlexNet are proved to be good in [19]. Features of multiple layers from AlexNet are used in object-oriented classification of remote sensing images, proving that fully connected layer features give good results when compared with convolution layers and are more expressive than spectral or texture features [20]. Klemen Grm in [21] conducted experimentation for face recognition application comparing various CNN models such as AlexNet, VGGNet, GoogLeNET, etc. Pre-trained AlexNet is used for speech emotion recognition representing speech features as images [22]. Similarly, NLP problems like automatic speech recognition for phonetic classification [23] are also solved with CNN. We have studied the performance of features extracted from fully connected layers of AlexNet when used for image registration.

The performance of most of the applications mentioned previously depends on the computation of similarity between the feature vectors. Conventional approaches using hand crafted features needed to find a suitable similarity measure depending on the application and type of images being employed. Table I lists some of the works reported. It can be observed that choosing suitable distance measure is necessary even when similar features are used across different applications. Hence when features extracted from pre-trained convolution layers of CNN's are employed choice of similarity measure affects the performance of applications. We have tested the effect of distance measure in image registration when features are extracted from convolution layers of AlexNet.

TABLE I. LIST OF THE FEATURES, DISTANCE MEASURES USED IN APPLICATIONS

| Reference | Application | Distance measures compared | Features Used |
|---|---|---|---|
| Yossi Rubner et al. [24] | Image retrieval | EMD, Jeffrey divergence, Chi-square statistic, L1-norm, Euclidean | Colour and texture |
| JesusAngulo, et al. [25] | Image classification and retrieval | Cityblock differences, Euclidean distance, Chi-square distance, Mahalanobis | Colour and texture |
| Ivan Laptev et al. [26] | Image matching | Squared difference of Gaussian function | Colour, blob and ridge |
| Xi Chen et al. [27] | Image matching | Cross affinity distance, SSD, Chamfer, Bhattacharyya | Edge orientation |
| Tudor Barbu et al. [28] | Image recognition | Euclidean | Colour |
| A. Melbourne et al.[29] | Image registration | Mutual information | Intensity |
| Xia et al. [30] | Satellite image indexing | Kullback-leibler Divergence | Shape |
| AbhjieetKumar Sinha et al. [31] | Image retrieval | Euclidean, Manhattan, Cosine | Colour |
| Abul Hasnat et a1. [32] | Face similarity | Modified Euclidean, Cosine Manhattan | Colour |
| Dengsheng Zhang et al. [33] | Image retrieval | Minkowski, L1-Norm, L2-Norm, chi-square statistic, Cosine, Mahalanobis, Quadratic, Histogram Intersection | Contour, region |

## III. FEATURE BASED IMAGE REGISTRATION

A classic pipeline for Feature based image registration contains the following steps:

*1) Feature detection*: To detect a pixel of an image as a feature key point, properties of every pixel (such as gradient) are examined with respect to its neighborhood. To improve the invariance to deformations, detection algorithms have used scale space and affine space of images. Scale invariant feature transform, SIFT [21], detects features at different scales. Difference of Gaussian is applied for a series of smoothed and resampled images to know the local extremas around the interested point which gives stable points further known as key points.

*2) Feature description:* Each key point is described by a vector of values. A window of size N x N is taken around a key point and given as input to AlexNet[8,11]. Activations from fc6 and fc7 are used as feature descriptors.

*3) Matching:* Key points of images are matched by computing the similarity between the corresponding feature descriptors. Different distance metrics such as Euclidian, Cosine, Manhattan etc. are used to compute the similarity. Different methods like 1-way nearest neighbor, 2-way nearest neighbor and their ratios are used to determine the correspondences.

*4) Transformation estimation:* The mapping function parameters are computed by establishing inliers among the feature correspondences. RANSAC is used to compute the homography matrix and the inliers.

### A. CNN based Feature Extraction using AlexNet

AlexNet with 5 Convolution layers, 5 relu layers, 5 max pooling layers, 2 fully connected layers or dense layers is used. Images are to be resized to 224x224 and given as input to AlexNet. Layer1, convolution layer convolves the input image 224x224x3 with 96 kernels of size 11x11x3 and with stride 4. Output of the layer is 55*55*96. Rectified linear unit activation and Max pooling functions are used to reduce over fitting and add nonlinearity to the extracted features, and the output is of size 27x27x96. Convolution Layer 2 which contains 128 kernels of size 5x5, stride 1x1, gives output 27x27x128. Similarly, convolution layers 3, 4 and 5 with 384, 384 and 256 filters, each of size 3x3 and stride 1x1 followed by max pool, dropout and padding are applied. Finally, the fully connected layer 6 and 7 gives 4096 features. Layer1 gives edge and blob of the input image, layer2 performs the conjunctions of these *edges or* responds to *corners* and other *edge* or color conjunctions*,* layer3 output is texture of a image, Layer5 identifies object parts, fc6 and fc7 gives image features.

### B. Distance Measures for Similarity

To find the most suitable distance measure between the images, a comparative study of features with dissimilarity measures is required. In this work, we consider dissimilarity measures like Euclidean, City block, Cosine, Minkowski, and Correlation through which we study the dissimilarity among the features of two images.

**Cityblock distance:** Measures the path between the pixels based on four connected neighbourhood.

$$Cityblock(p_k, q_k) = \sum_{k=1}^{n} (|p_k| - |q_k|)$$

**Euclidean distance:** Most commonly used metric to find the difference, calculates the square root of the sum of the absolute differences between two feature points.

$$Euclidean(p_k, q_k) = \sqrt{\sum_{k=1}^{n} (p_k - q_k)^2}$$

**Cosine distance:** Finds the normalized dot product of the two feature points.

$$Cosine(p_k, q_k) = \sum_{k=1}^{n} \frac{(p_k \cdot q_k)}{||p_k|| \cdot ||q_k||}$$

, where • indicates vector dot product

**Minkowski distance:** Is a generalization of Euclidean Distance.

$$Minkowski(p_k, q_k) = \left( \sum_{k=1}^{n} |p_k - q_k|^r \right)^{1/r}$$

**Correlation distance:** The *correlation* of feature two points, p and q, with *k* dimensions is calculated as:

$$Correlation(p_k, q_k) = \sum_{k=1}^{n} \frac{Cov(p_k \cdot q_k)}{Std(p_k) \cdot Std(q_k)}$$

$$Cov((p_k, q_k)) = \frac{1}{n} \sum_{k=1}^{n} (p_k - \bar{p}) \cdot (q_k - \bar{q}),$$

$$Std((p_k)) = \frac{1}{n} \sum_{k=1}^{n} (p_k - \bar{p}), \bar{p} = \frac{1}{n} \sum_{k=1}^{n} (p_k)$$

## IV. EXPERIMENTATION

The performance of various distances measures on fc6 and fc7 outputs of AlexNet, for the purpose of image matching and registration is objectively evaluated using benchmark dataset. Our implementation starts with a) detecting features using SIFT, b) describing and extracting features from fc6 and fc7 layers of AlexNet, c) finding differences between the features points by using different dissimilarity measures, d) finding matches using various matching algorithms like nearest neighbour (NN), nearest neighbour ratio (NNR), with one way and two-way matching. Matching performance is noted for various threshold values of 0.3, 0.5 and 0.7 for NN, 1.1, 1.2 and 1.3 for NNR, e) estimating homography matrix from the matches using RANSAC, f) computing the various evaluation measures described in the following section. All the above implementation is done in Matlab and executed on i7 CPU@2.7GHz with 8GB RAM.

### A. Dataset

The dataset is freely available at http://www.robots.ox.ac.uk /~vgg/data/data-aff.html which contains 8 subsets of images. Each such subset contains six images, with first image being the original image and the rest of 5 images having different effects like zoom, rotation, illumination, compression, view angle etc. sample images are shown in Fig. 1.

### B. Evaluation Measures

Evaluation is conducted at three stages, to find a relevant distance measure among the tested measures such as Euclidean, Cosine etc for the AlexNet features, relevant matching method between NN and NNR, and the suitable features between fc6 and fc7.

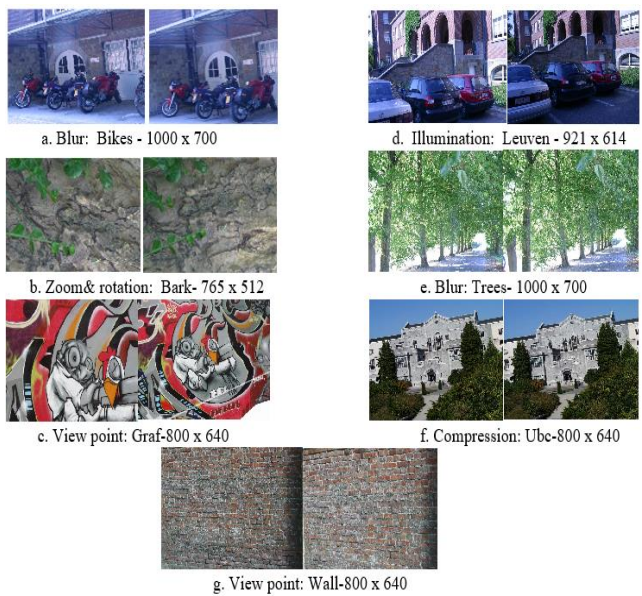

Fig. 1. Sample Images from Graffiti Dataset.

Evaluation is done using ground truth homography matrix by computing the following values:

Keypoint error with ground truth (KE_GH): The distance between the first image keypoint matches which are transformed with ground truth homography matrix in to second image and second image keypoint matches.

True Positive matches (TP): number of correct matches that are found with distance less than 2 pixels when compared with ground truth matches.

Using computed homography matrix (from RANSAC) the following values are computed:

Keypoint error with computed homography (KE_CH): The Distance between the first image keypoint which are transformed with computed homography matrix in to second image and second image keypoint matches. Computed homography matrix is estimated by applying RANSAC.

Inlier Ratio (IR): ratio of the total number of inliers and the total number of matched key points.

Higher the value of TP and IR better the matching whereas low values of KE_CH, KE_GH indicate that registration accuracy is better.

## V. PERFORMANCE AND RESULTS

In this segment of the paper we present results obtained and discuss the performance at various evaluation stages. Results are interpreted by comparing images with different deformations (like scale, rotation, zoom, blur, illumination, compression).

### A. Suitable Distance Measure

In first stage we matched the feature vector of image1 with the other image feature vector by using dissimilarity measures. Various distance measures considered detect correspondences using different thresholds in one/two-way NN and NNR matching techniques.

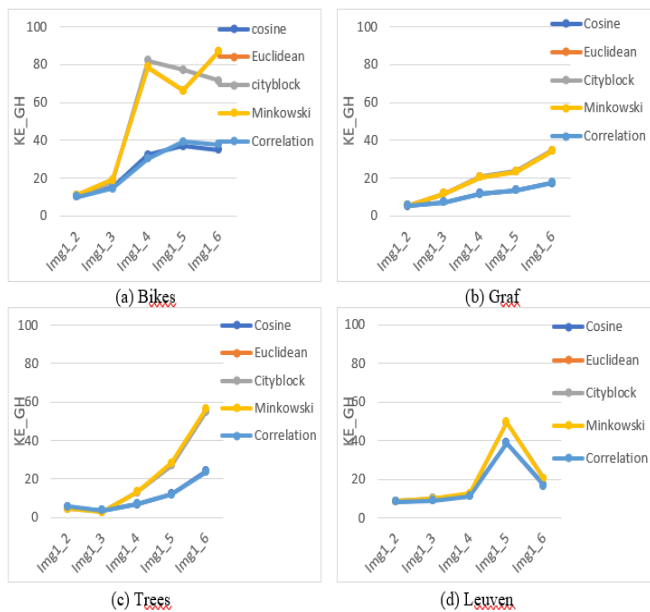

Fig. 2. Key Point Error On Ground Truth Homography for all 5 Images when Matched with First Image using Features of fc6 Layers, 1NNR with Threshold 1.1 for Various Dissimilarity Measures.

The key point error rate between correspondences of the image 1 and image 2 is computed with given ground truth homography matrix and computed homography.

We present the results of 4 different images in Fig. 2. From graphs it can be observed that KE_GH is less between first two images as compared between any other image pair. With any kind of deformation present in the dataset the most similar image is image2. It can be observed that Cosine and Correlation give better performance across deformations when compared to other distance measures.

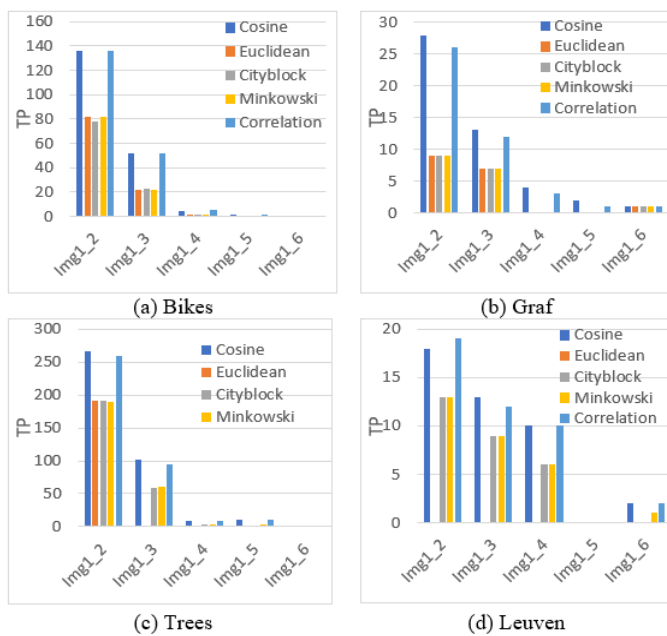

Fig. 3. True Positive Matches for all 5 Images when Matched with First Image using 1NNR with Threshold 1.1 and the Features of fc6 Layers Across 5 Dissimilarity Measures.

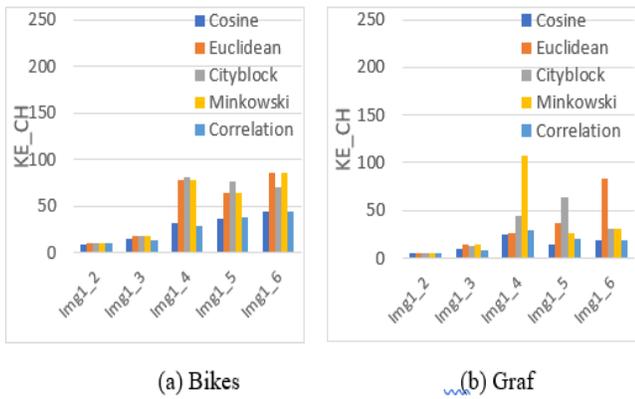

Fig. 4. Key Point Error for Computed Homography for 5 Images when Matched with First Image using 1NNR with Threshold 1.1 and the Features of fc6 Layers Across 5 Dissimilarity Measures.

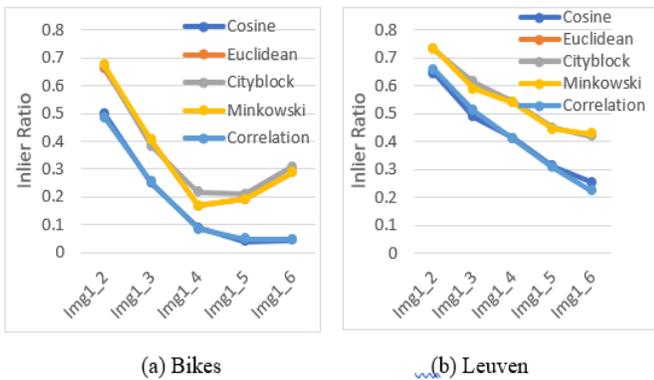

Fig. 5. Inlier Ratio for 5 Images when Matched with First Image using 1NNR with Threshold 1.1 and the Features of fc6 Layers Across 5 Dissimilarity Measure.

Similarly, number of matches, key point error for the computed homography and inlier ratio are presented in Fig. 3, Fig. 4 and Fig. 5 respectively. We used 1-way NNR with threshold 1.1 as the matching technique (we experimented with various thresholds of 1.1, 1.2, 1.3), to find the best dissimilarity measure. From the graphs it can be observed that error rate with respect to both ground truth and computed homography is less in the case of cosine and correlation distance measures. At the same time, we can see that the number of true positives is more (Fig. 3) for all the types of images with cosine and correlation, which are dominating other distance metrics. Cosine is the measure which is consistently performing well, when compared to any other dissimilarity measure.

*B. Suitable Matching Technique*

We tried to establish the best matching technique based on dissimilarity measure which is found to be the better in the above scenario.

From the results in Fig. 6 it can be observed that 1way NNR and 2way NN are rational, when compared to 1NN and 2 NNR. In some cases, 2 NNR shows high values of inlier ratio, but is not efficient as the number of matches is very less (< 10). Matches obtained for one image pair are shown in Fig.7 and Fig. 8.

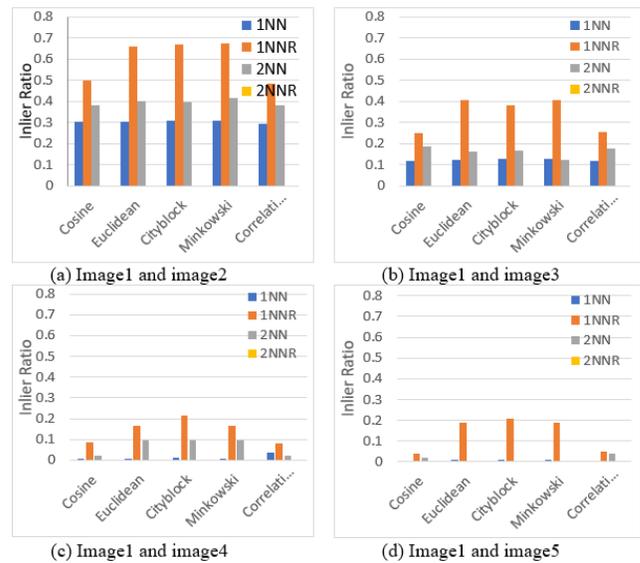

Fig. 6. Inlier Ratio for 4 Images (from 2 to 5) when Matched with first Image of Bikes using 1NN, 2NN with Threshold 0.5 and 1NNR, 2NNR with Threshold 1.1, for the Features of fc6 Layers Across 5 Dissimilarity Measures.

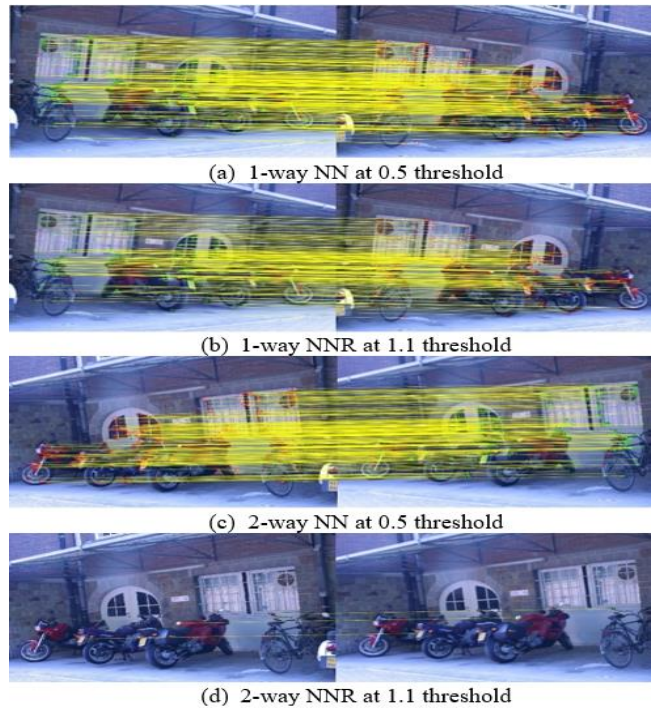

Fig. 7. Matches with Different Matching Techniques of first Image to Second Image of Bikes, with fc6 Features and Cosine Dissimilarity.

It is clearly noticed that the number of correspondences are better in 1-way NN compared to any other matching technique. However, the key point error is less in the case of 2-way NNR as the matched points are accurate when compared to any other matching technique. However, 1-way NNR and 2-way NN error rate is moderate and these matching methods are good across all images of dataset such as ubc, graf, leuven trees and wall. Hence it can be concluded that 1way NNR is best and 2 NN second from matching results of the experiments.

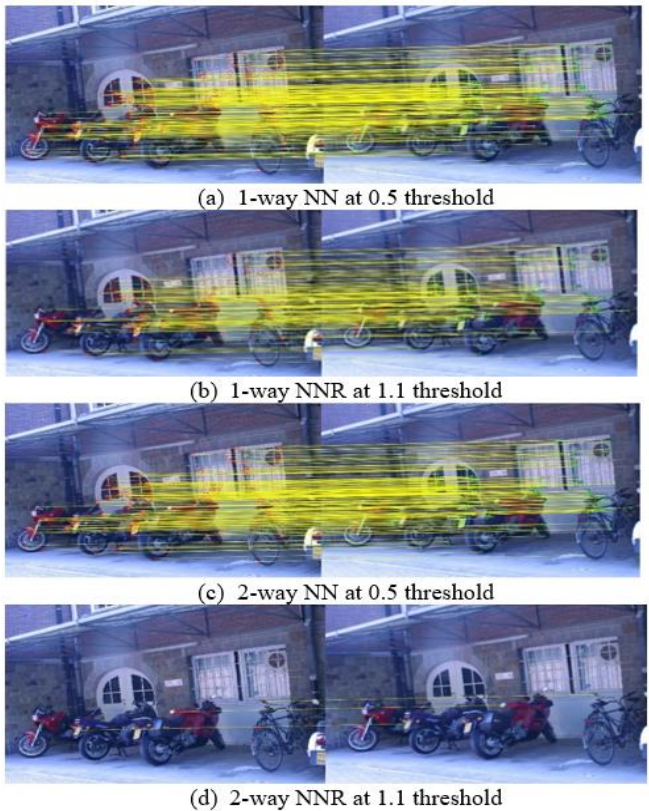

Fig. 8. Matches with Different Matching Techniques of First Image to Second Image of Bikes, with fc7 Features and Cosine Dissimilarity.

### C. Best Image Features among Fc6 and Fc7

Finally, we test the features of the two layers, fc6 and fc7 of AlexNet to find the better features to be used for Image registration. Based on the above experiments, we present cosine as dissimilarity measure for matching first and second images and present the results in Fig. 9.

Finally, in addition to the dissimilarity and matching measures, we found that instead of fc7 features fc6 layer features are giving quantitatively more matches for the same distance measure consistently.

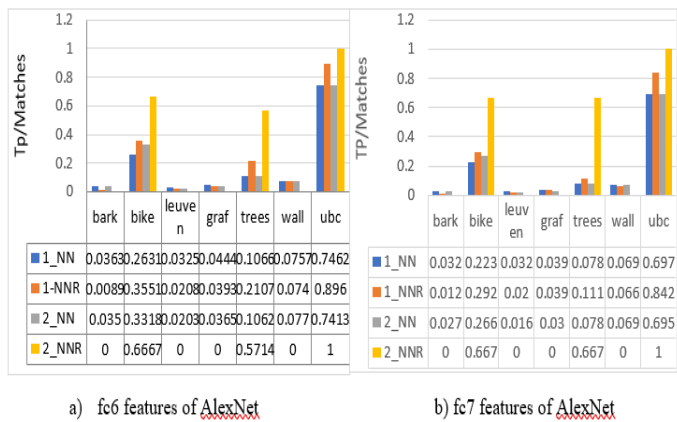

Fig. 9. True Positive Matches between first and Second Image Across All Matching Techniques with Cosine as a Dissimilarity Measure.

### VI. CONCLUSION

Trained CNN model, AlexNet, is used as feature extractor for registering images with variations such as zoom, rotation, lighting etc. Outputs of fully connected layers, fc6 and fc7 are used as feature descriptors by giving as input a region around the key point of image, which is detected using SIFT. In order to obtain good registration results, evaluation of various distance measures and matching methods is performed. Objective evaluation measures computed from ground truth are used to compare matching and registration performance. It has been observed that Cosine dissimilarity measure, followed by correlation, consistently gives better matching and registration across images of various deformations. Among the various matching strategies tested, results from one way nearest neighbour ratio with a threshold of 1.1 and two way nearest neighbour with a threshold of 0.8 are promising. Our future work involves verifying the effect of distance measures with other CNNs such as VGG and further to design a deepnet to learn similarity between image features.

REFERENCES

[1] Barbara Zitova, Jan Flusser, Image Registraion methods: a survey, Elsevier, Image and Vision Computing, 21, pp. 977-1000, 2003.

[2] David G. Lowe, Distinctive Image Features from Scale-invariant keypoints, Internationsl journal of Computer Vision, Vol. 50, no.2, pp91-110, 2004.

[3] Herbert Bay, Andreas Ess, Tinne Turtelaars, and Luc Van Gool, SURF: Speeded Up Robost Features, Computer Vision and Image Understanding (CVIU), Vol. 110, No. 3, pp. 346-359, 2008.

[4] J. Matas, O. Chum, M. Urban, T. Pajdla, Robust Wide Baseline Stereo from Maximally Stable Extremal Regions, Proc. of British Machine Vision Conference (BMVC), pp. 384-396,2002.

[5] Martin A. Fischler and Robert C. Bolles, SRI International, Graphics and Image Processing, Communications of ACM, vol.24, 1981.

[6] Ebrahim Karami, Siva Prasad, Mohamed Shehata, Image Matching Using SIFT, SURF, BRIEF, and ORB, Performance Comparison for Distorted Images, in proc. of the 2015 newfoundland Electrical and Computer Engineering Conf. Canada, Nov 2015.

[7] M. Hassaballah, Aly Amin Abdelmgeid and Hammam A. Alshazly, Image features Detection, description and Matching, Spring Internation Publishing Switzerland, DOI 10.10007/978-3-319-28854-3_2, 2016.

[8] Krizhevsky, Alex, Ilya Sutskever, and Geoffrey E. Hinton. "ImageNet Classification with Deep Convolutional Neural Networks." *Advances in neural information processing systems*. 2012.

[9] Zheng-Wu Yuan, Jun Zhang, Feature Extraction and Image retrieval based on AlexNet, 8[th] ICDIP, china, 2016.

[10] Karen Simonyan and Andrew Zisserman, Very Deep Convolution Networks for large-Scale Image Recognition, published as a Conf. paper at ICLR 2105.

[11] Md. ZahangirAlom, Tarek M.Taha, Christopher Yakopcic, Stefan Westberg, Moahudul Hasan, Brain C Van Esesn, Abdul A. S. Awwal, and Vijayan K.Asari, The History Began fronAlexNet: A Comprensive Survey on DeepLearning Approaches, arXiv:1803.01164, 2018.

[12] Krizhevsky, Alex, Ilya Sutskever, and Geoffrey E. Hinton. "ImageNet Classification with Deep Convolutional Neural Networks." Communication of the *ACM, Vol 6, pp.84-90,* 2017.

[13] Yann LeCun, Leon Bottou, YoshuaBengio, and Patrick Haffner, Gradient -Based Learning Applied to Document Recognition, Proc. Of the IEEE, 1998.

[14] Christian Szegedy, Wei Liu, Yangqing Jia, Pierre Sermanet, Scott Reed, Dragmir Anguelov, Dumitru Erhan, Vincent Vanhoucke, Andrew Rabinovich, Going deeper with convolutions, IEEE conference on Computer Vision and Pattern Recognition, DOI: 10.1109/CVPR.2015.7298594, pp.1-9,2015.

[15] Yiren Zhou, Hossein Nejati, Thanh-ToanDo, Nagai-Man Cheung, Lynette Cheah, "Image-based Vehicle analysis using Deep Neural network: A systematic Study", IEEE International conf. on digital signal processing, 2016.

[16] Lisha Xiao, Qin Yan, Shuyu Deng, Scene Classification with improved AlexNet Model, 12th international conf. on intelligent systems and knowledge engineering, 2017.

[17] Marmanis. D, Datu .M, Esch.T, Stilla. U, Deeplearning earth observation classification using ImageNet Pretrained Networks, IEEE Geo-Science Remote Sensing, 2015.

[18] Jing Sun, Xibiao Cai, Fuming Sun, Jianguo Zhang, Scene Image Classification method based on Alex-Net Model, 3rd international comf. on informative and cybernetics for computational systems, Aug 2016.

[19] Hailong Liu, Baoan Li, XueqiangLv, Yue Huang, Image Retrieval Using Fused Deep Convoluiton features, International congress of Inforamtion and Communication Technology, procedia computer science 107, pp. 749- 754, 2017.

[20] Ling Ding, Hongyi Li, Changmiao Hu, Wei Zhang, Shumin Wang, AlexNet Feature Extraction and Multi-kernel Learning for Object Oriented classification, symposium "Developments, technologies and Applications in Remote sensing", internatonal Archives of Photogrammetry, Remote Sensing and Spatial Information Sciences, Vol. XLII-3, China, may 2018.

[21] KlemenGrm, VitomirStruc, Anais Aritges, Matthieu Caron, Hazim Kemal Ekenel, Strengths and weaknesses of Deep Learning Models for Face Recognition Against Image Degradations, Submitted for publication to IET Biometrics, oct 2017.

[22] Margaret Lech, Robert S. Bolia, Real Time Speech Emotion Using RGB Image Classification and Transfer Learning, DOI, 10.1109/ICSPCS 2017.8270472, dec 2017.

[23] Hinton. G, Li Deng, Dong Yu, George Dahl, et al. Deep neural networks for acoustic modelling in speech recognition, IEEE signal processing Magazine 29, 82-97, 2012.

[24] Yossi Rubner, Carlo Tomasi and Leonidas J. Guibas, The earth mover's distance as a metric for image retrieval, International Journal of Computer Vision 40(2), pp 99-121,2000.

[25] Jeaus Angulo and Jean Serra, Morphological Color size distributions for image classification and retrieval, Proc. Of Advanced Concepts for Intelligence Vision Systems, 2002.

[26] Ivan Laptev and Tony Lindeberg, A distance measure and a feature likelihood map concept for scale-invariant matching, International Journal of Computer Vision 52(2/3), pp 97-120,2003.

[27] Xi Chen and Tat-Jen Chem, Learning feature distance measures for image correspondences, IEEE Conf. on Computer Vision and Pattern Recognition, San Diego, CA, 2005.

[28] Tudor Barbu, Adrian Ciobanu, and Mihale Costin, Automatic color-based image recognition technique using LAB-features and robust unsupervised clustering algorithm Advances in Information Science, Circuits and Systems ISBN:978-1-61604-009-2, 2007.

[29] A. Melbourne, G. Ridgway and D. J. Hawkes, Image similarity metrics in image registration, Proceedings of SPIE- the International Society for Optical Engineering, DOI:10.11117/12/840389, 2010.

[30] Xia, G.S.; Yang.W; Delon J; Gousseau . Y; Sun .H; Maitre .H, Structural high-resolution satellite image indexing, In Proc. of the ISPRS TC VII Sym. -100 years, Vienna, Austria, pp. 289-303, 2010.

[31] Abhijeet Kumar Sinha, K. K. Shukla, A study of distance metrics in histogram-based image retrieval, International Journal of Computers and Technology, Council of innovative Research, pp 821-830, 2013.

[32] Abul Hasnat, Santanu Halder, D. Bhattacharjee, M. Nasipuri and D. K. Basu, Comparative study of distance metrics for finding skin color similarity of two color facial images, 2013.

[33] Dengsheng Zhang and Guojun Lu, Evaluation of Similarity Measurement for Image Retrieval, International Conference on Neural Networks and Signal Processing, DOI: 10.1109/ICNNSP.2003.1280752, 2004.